\documentclass[11pt,a4paper]{article}
\usepackage[hyperref]{acl2019}
\usepackage{times}
\usepackage{latexsym}
\usepackage[OT2,OT1]{fontenc}
\newcommand\cyr
{
\renewcommand\rmdefault{wncyr}
\renewcommand\sfdefault{wncyss}
\renewcommand\encodingdefault{OT2}
\normalfont
\selectfont
}
\DeclareTextFontCommand{\textcyr}{\cyr}

\usepackage{url}

\usepackage[utf8]{inputenc}

\usepackage{amssymb}
\usepackage{amsmath}
\usepackage{mathtools}
\usepackage{hhline}
\usepackage{comment}

\DeclareMathOperator*{\argmax}{arg\,max}

\aclfinalcopy 

\title{Unsupervised Bilingual Lexicon Induction Across Writing Systems}

\author{Parker Riley and Daniel Gildea \\
  Department of Computer Science\\ University of Rochester\\Rochester, NY 14627 \\
  }

\date{}

\begin{document}
\maketitle
\begin{abstract}
  Recent embedding-based methods in unsupervised bilingual lexicon induction have shown good results, but generally have not leveraged orthographic (spelling) information, which can be helpful for pairs of related languages. This work augments a state-of-the-art method with orthographic features, and extends prior work in this space by proposing methods that can learn and utilize orthographic correspondences even between languages with different scripts. We demonstrate this by experimenting on three language pairs with different scripts and varying degrees of lexical similarity.
\end{abstract}

\section{Introduction}
\vspace{-0.2cm}
Bilingual lexicon induction is the task of creating a dictionary of single-word translations between two languages. Recent approaches have focused on learning crosslingual word embeddings, where translations are close in a shared vector space \citep{VulicMoens:2013,mikolov2013exploiting,artetxe-2016}. More recently, unsupervised methods have been developed that are applicable to low-resource pairs \citep{artetxe-2017,zhang-EtAl:2017,Lample2018word}. These have the benefit of producing embeddings that enable unsupervised machine translation \citep{artetxe2018,Lample2018mt}.

These models generally ignore lexical features such as spelling. However, prior work used string edit distance and subword units as features for similar tasks \citep{Dyer:ACL11,Berg:10,haghighi-acl08}. These features can be useful for related languages such as English and German (consider the pair \textit{doctor}-\textit{Doktor}).

\citet{Riley18} explored using orthographic information for embedding-based lexicon induction, but they assumed a shared alphabet, which is less useful for related languages that use different scripts (such as Hindi and Bengali). Here, we combine the core approach of \citet{artetxe2018robust} with three extensions of the orthography-based methods of \citet{Riley18}, and evaluate them on three language pairs with different scripts and varying degrees of lexical similarity and available resources: English-Russian, Polish-Russian, and Hindi-Bengali.

\section{Background}
\vspace{-0.2cm}
This work is based on the framework of \citet{artetxe2018robust}. Following their work, let $X \in \mathbb{R}^{|V_X| \times d}$ and $Z \in \mathbb{R}^{|V_Z| \times d}$ be the word embedding matrices of a source and target language, such that each row corresponds to a word's $d$-dimensional embedding. The $i$th row of one of these matrices is $X_{i*}$ or $Z_{i*}$. The vocabularies are $V_X$ and $V_Z$.

Each embedding dimension is mean-centered, and each embedding is length-normalized.

The goal is to find two linear transformation matrices $W_X$ and $W_Z$ that project the two languages' embeddings into a shared vector space. Given an initial dictionary $D \in \{0,1\}^{|V_X| \times |V_Z|}$ where $D_{ij} = 1$ if target word $j$ is a translation of source word $i$, we wish to find:
\vspace{-0.1cm}
$$\argmax_{W_X,W_Z} \sum\limits_i\sum\limits_j D_{ij}((X_{i*}W_X) \cdot (Z_{j*}W_Z))$$

\citet{artetxe2018robust} note that the optimal solution to this is $W_X = U$ and $W_Z = V$, given the singular value decomposition $U\Sigma V^\intercal = X^\intercal DZ$.

A new dictionary $D^{'}$ is produced from the similarity matrix $S = XW_XW_Z^\intercal Z^\intercal$ such that $D^{'}_{ij} = 1$ if $j = \argmax_k (X_{i*}W_X) \cdot (Z_{k*}W_Z)$. $D^{'}$ is used for the next iteration of the process, and this repeats until convergence.

This process is guaranteed to converge to a local optimum of the objective, but the quality of the initial dictionary $D$ is important, and \citet{artetxe2018robust} found that a random initialization generally results in poor performance. They proposed a fully unsupervised dictionary initialization method that assumes approximate isometry between the two embedding spaces.

Each iteration, elements in $S$ are set to $0$ with probability $1-p$, where $p$ grows to $1$ over time, to encourage exploration early in training. Also, only the top $20,000$ most frequent words in each language are used for training, to limit overhead. $S_{ij}$ is discounted by the average cosine similarity between each word and its $10$ nearest neighbors in the other language. Finally, the dictionary is constructed bidirectionally, considering each source word's nearest target word and vice versa, so entries in $D$ can be $2$ for mutual nearest neighbors.

Once the keep probability $p$ has grown to $1$ and the objective value does not increase within $50$ iterations, a modified final iteration is performed where $X$ and $Z$ undergo a ``whitening'' transformation, giving each dimension unit variance and $0$ covariance with other dimensions. This is reversed after calculating the mapping matrices.

We propose three modifications to this system, described in the following section.

\section{Incorporating Orthography}
\vspace{-0.2cm}
The motivation for the three modifications proposed here is that the words used for a given concept are not in general independent across languages: many pairs of languages feature words that were inherited from a common ancestor language (consider the prevalence of Latin roots in many European languages) or that were borrowed from the same third language. These words often undergo changes in pronunciation and spelling when entering the new language, but we believe that in many cases there is enough remaining surface similarity that a bilingual lexicon induction system can benefit from it. This has been demonstrated for languages with mostly similar alphabets \citep{naim-decipherment18,Riley18}, but in this work we seek to develop methods that are applicable to languages with different alphabets: consider the related languages Polish and Russian and their phonetically similar translation pair \textit{trudna}-\hspace{-0.4cm}{\cyr trudno}.
\subsection{Orthographic Embedding Extension}\label{oe}
This method expands the embeddings of source and target words with additional dimensions that correspond to scaled character unigram and bigram counts within the respective word. The motivation is that if we imagine a pair of languages $\alpha$ and $\beta$ where words in $\alpha$ were transliterated from $\beta$ using a character substitution cipher, then there exists a linear transformation that can be applied to the vectors of character counts for the words in $\alpha$ to yield the corresponding count vectors in language $\beta$. The mapping matrices $W_X$ and $W_Z$ can encode this transformation, as well as an interaction between the count vectors and the core embeddings. 

Mathematically, let $A$ be an ordered set of character $n$-grams (an alphabet), containing the top $k = 100$ most frequent characters and most frequent character bigrams from each language.

Let $O_{X}$ and $O_{Z}$ be the orthographic extension matrices for each language, containing counts of the character unigrams and bigrams appearing in each word $w_i$, scaled by a constant factor $c$:
$$O_{ij} = c \cdot \mathrm{count}(A_j, w_i), O \in \{O_{X}, O_{Z}\}$$

These extension matrices are appended to $X$ and $Z$, such that $X^{'} = [X ; O_{X}]$ and $ Z^{'} = [Z ; O_{Z}]$

$X^{'}$ and $Z^{'}$ are length-normalized and mean-centered as normal. The rest of the training loop is unmodified, with one exception: the whitening transformation applied before the final iteration, which removes the correlation between the embedding dimensions, does not make sense for the dimensions added by this method because many of the character $n$-gram counts are dependent on one another. This is trivially true for character bigrams in $A$ that are composed of unigrams also in $A$. Therefore, we remove the additional dimensions immediately before the final iteration. We observed that skipping this step results in a catastrophic loss in performance ($<1\%$ accuracy).

\subsection{Learned Edit Distance}
The previous method uses a bag-of-character-$n$-grams approach, but this method considers both words' character sequences to learn an edit distance and use it to modify $S$.
The standard Levenshtein string edit distance with uniform substitution costs \citep{Levenshtein} is uninformative with disjoint alphabets, so we use the method of \citet{Ristad:1998} to learn an edit distance function.

This method models $p(x_1^N,z_1^N | \theta)$, where $x_1^N$ is a source word of length $N$ and $z_1^M$ is a target word of length $M$. This is the probability of generating the words with a sequence of operations $\langle a_X, a_Z \rangle$ where $a_X \in A_X^{'} \cup \{\epsilon\}$ and $a_Z \in A_Z^{'} \cup \{\epsilon\}$ are character $n$-grams in the source and target alphabets, respectively, and the concatenation of the operations yields the string pair. These alphabets are similar to those in Section~\ref{oe}, except they include \textit{all} character unigrams, and the same number of bigrams, to maximize coverage. $\epsilon$ is the zero-length string; the null operation $\langle \epsilon, \epsilon\rangle$ is disallowed. 

The parameters $\theta$ consist of per-operation probabilities $\theta(a_X, a_Z)$. $p(x_1^N,z_1^M | \theta)$ can be calculated using dynamic programming, where each table entry $\alpha(n,m)$ is the forward probability of generating the prefix pair $(x_1^n, z_1^m)$:
\begin{multline*}
\alpha(n,m) = \\
\sum_{j=0}^{J}\sum_{k=I(j=0)}^{K}{\theta(x_{n-j+1}^n,z_{m-k+1}^m)\alpha(n-j,m-k)}
\end{multline*}
where $J$ and $K$ are the maximum lengths of character $n$-grams in the source and target alphabets, $I(j=0)$ is $1$ if $j=0$ and $0$ otherwise, $x_{i+1}^i = \epsilon$, and $\alpha(0,0) = 1$. The probability is calculated as $p(x_1^N,z_1^M | \theta) = \alpha(N,M)$.

The parameters $\theta$ are learned using the Expectation Maximization algorithm \citep{Dem77}. The algorithm features $\beta$, a backward counterpart to $\alpha$ such that $\beta(n,m)$ is the probability of generating the suffix pair $(x_{n+1}^N,z_{m+1}^M)$. These are used to calculate expected counts for all operations $\langle a_X, a_Z \rangle$ and then update the parameters $\theta(a_X,a_Z)$.

Because we are interested in bilingual lexicon induction methods that require minimal supervision, we use synthetic training data to learn $\theta$. This data is produced by running the unmodified system of \citet{artetxe2018robust} and recording the 5,000 highest-similarity word pairs. We run 3 iterations of EM on this data to learn $\theta$. We then modify the entries in the similarity matrix $S_{ij}$ at each iteration as follows:
$$S_{ij} \mathrel{+}= c \cdot \max(0,1 + \frac{\frac{\log{p(w_i,w_j |\theta)}}{\max(|w_i|,|w_j|)}}{\log{((1+|A^{'}_X|)(1+|A^{'}_Z|))}})$$

This equation boosts the similarity of word pairs with a per-operation log-probability that is higher than chance.

To avoid calculating the score for all word pairs, we only evaluate a subset, identified using an adaptation of the Symmetric Delete spelling correction algorithm described by \citet{Garbe2012}. This algorithm runs in linear time with respect to the vocabulary size and identifies all word pairs that are identical after no more than $k$ character deletions from each word; we use $k = 2$.

To adapt this algorithm to our context of languages with disjoint alphabets, we first use our string edit probability model to transliterate each source word $x_1^N$ as $\argmax_{z_1^M} p(x_1^N,z_1^M|\theta)$. This is also done via dynamic programming, where each table entry $\delta(n)$ contains the maximum log-probability of any valid segmentation of the source prefix $x_1^n$ into elements of $A_X^{'}$, paired with the target sequence produced by substituting each with its max-probability element of $A_Z^{'}$. Mathematically, $\delta(n)$ (initialized to $-\infty$) is defined as:
$$\max_{j=1}^{J}{\max_{a_Z \in A_Z^{'} \cup \{\epsilon\}}{\log{(\theta(x_{n-j+1}^n,a_Z))}}+\delta(n-j)}$$
where $j$ ranges over lengths of source alphabet items. Because there are infinitely many target character sequences that could be produced by repeatedly substituting $\epsilon$ in the source string, we only consider segmentations of $x_1^N$ into non-empty subsequences.
\begin{table*}
\begin{center}
{\small
\begin{tabular}{ || c || c | c | c ||}
  \hline
  Method (hyperparameter selection) & English-Russian & Polish-Russian & Hindi-Bengali \\
  \hhline{||=||=|=|=||}
 	\citet{artetxe2018robust} & 44.20 & 49.70 & 23.50\\
  \hline
    Embedding extension (best dev accuracy) & 44.36 ($c = 0.2$) & 49.89 ($c = 0.3$) & \textbf{25.53} ($c = 0.15$) \\
  \hline
  	Embedding extension (best objective) & 43.12 ($c = 0.25$) & \textbf{50.01} ($c = 0.25$) & 25.29 ($c = 0.3$) \\
  \hline
  	Learned edit distance (best dev accuracy) & 44.91 ($c = 0.25$) & 49.98 ($c = 1.1$) & 25.31 ($c = 0.9$) \\
  \hline
  	Learned edit distance (best objective) & \textbf{45.05} ($c = 0.3$) & 50.00 ($c = 0.9$) & 25.31 ($c = 0.9$) \\
  \hline
  	Character RNN (best dev accuracy) & 44.77 ($c = 0.15$) & 49.94 ($c = 0.6$) & 24.99 ($c = 0.1$) \\
  \hline
  	Character RNN (best objective) & 44.612 ($c = 0.3$) & 49.94 ($c = 0.6$) & 24.33 ($c = 1.3$) \\
  \hline
  
\end{tabular}}
\end{center}
\caption{Test accuracies, averaged over 10 runs. The hyperparameter $c$ was selected either by development accuracy or objective value. Selection was based on a $5$-run average across $18$ values ranging from $0.05$ to $1.4$.}
\label{table:test}
\end{table*}
\subsection{Character RNN}
This method is similar to the previous method, except that instead of a string edit probability model, we use a bidirectional recurrent neural network to estimate the orthographic similarity of two strings. We use a publicly-available sequence-to-sequence (seq2seq) library from IBM.\footnote{https://github.com/IBM/pytorch-seq2seq} It uses an encoder-decoder architecture, where the encoder is bidirectional and has 128 hidden nodes and the decoder has $256$ hidden nodes, an attention mechanism, and a dropout probability of $0.2$. Both components use Gated Recurrent Units \citep{cho:2014}.

As before, we train the model on word pairs identified by the unmodified embedding mapping system, though we use $50,000$ pairs as training data and $20,000$ as development data, optimized by Adam \citep{kingma2014adam}.

After training, the RNN can be used to calculate the probability $p_{RNN}(w_j|w_i)$ of a target sequence $w_j$ given a source sequence $w_i$. This probability is used similarly to the edit probability from the previous section to update the similarity matrix $S$, with some modifications:
$$S_{ij} \mathrel{+}= c \cdot \max(0,1 + \frac{\frac{\log{p_{RNN}(w_j|w_i)}}{|w_j|}}{\log{(|A^{''}_Z|))}})$$
where $A_Z^{''}$ is the set of target character unigrams observed in the training data. 

\section{Experiments}
\vspace{-0.2cm}
We use data sets and pre-trained word embeddings available from the Facebook AI Research MUSE library\footnote{https://github.com/facebookresearch/MUSE} in our experiments. We evaluate our method on three language pairs: English-Russian, Polish-Russian, and Hindi-Bengali.

All embeddings are 300-dimensional and were trained on the language's section of the Wikipedia corpus using fastText \citep{Bojanowski17}.

Our methods do not require a training dictionary, but they do feature a hyperparameter $c$ that controls the relevance of the orthographic signal. We experiment with two methods of selecting $c$: accuracy on a held-out development set of $5,000$ source-target pairs, and the value of the objective (average similarity) over the entire vocabulary. The latter has the benefit of not requiring any data, allowing these methods to be fully unsupervised.

Each source word may have multiple correct translations, and predicting any yields full credit. The average number of translations per source word was approximately 1.3 for English-Russian, 1.4 for Polish-Russian, and 1.5 for Hindi-Bengali. All test dictionaries consist of $5,000$ pairs. 

The English-Russian dictionary was created from the one available in the MUSE library. However, the other two language pairs did not have their own dictionaries, so we created a dictionary by using English as a pivot language. To produce an $X$-to-$Z$ dictionary $D^{X \rightarrow Z}$ from $D^{X \rightarrow E}$ and $D^{E \rightarrow Z}$, we set $D^{X \rightarrow Z}_{ij}$ to $1$ if there exists a pivot word $k$ such that $D^{E \rightarrow Z}_{kj} = 1$, $D^{X \rightarrow E}_{ik} = 1$, and there is only $1$ word $k^{'}$ such that $D^{X \rightarrow E}_{ik^{'}} = 1$. This final constraint limits the number of translations for the source words, and limits the errors inherent to inferring a dictionary in this way.
\section{Results and Discussion}
\vspace{-0.2cm}
Table~\ref{table:test} shows the results of our experiments on test data. We identified the best hyperparamter $c$ from a range of possible values for each of our methods using development accuracy and the objective value, and report test accuracies for both methods, averaged over 10 runs. 

We see that the proposed methods universally outperform the baseline with the correct choice of $c$. The optimal scaling constant for Polish-Russian for each of the three methods is higher than for English-Russian, which meets our expectations given that the former are more similar.

The performance gain for Hindi-Bengali is considerably larger than for the other two, and the raw accuracy is much lower across all models; we hypothesize that this is because the Hindi and Bengali embeddings are of lower quality, because the corresponding Wikipedia data sets are roughly one tenth the size of the Polish and Russian sets, which are themselves roughly one tenth the size of the English set. This diminished quality hinders the underlying embedding mapping framework. However, this illustrates that orthographic information is more beneficial for low-resource languages, and the low-resource context has the best motivation for using unsupervised methods in the first place.
\section{Conclusion and Future Work}
\vspace{-0.2cm}
In this work, we presented three techniques for using orthographic information to improve bilingual lexicon induction for related languages with different alphabets. These techniques are applicable to low-resource language pairs because they do not require a human-annotated dictionary.

For future work, we are interested in extending these methods to unsupervised machine translation and developing a method for estimating optimal scaling constants directly, without needing to guess-and-check many possibilities.

\bibliography{all}
\bibliographystyle{acl_natbib}
\end{document}